# Noun-Phrase Analysis in Unrestricted Text for Information Retrieval

David A. Evans, Chengxiang Zhai

Laboratory for Computational Linguistics

Carnegie Mellon Univeristy

Pittsburgh, PA 15213

dae@cmu.edu, cz25@andrew.cmu.edu

## Abstract

Information retrieval is an important application area of natural-language processing where one encounters the genuine challenge of processing large quantities of unrestricted natural-language text. This paper reports on the application of a few simple, yet robust and efficient noun-phrase analysis techniques to create better indexing phrases for information retrieval. In particular, we describe a hybrid approach to the extraction of meaningful (continuous or discontinuous) subcompounds from complex noun phrases using both corpus statistics and linguistic heuristics. Results of experiments show that indexing based on such extracted subcompounds improves both recall and precision in an information retrieval system. The noun-phrase analysis techniques are also potentially useful for book indexing and automatic thesaurus extraction.

## 1 Introduction

### 1.1 Information Retrieval

Information retrieval (IR) is an important application area of natural-language processing (NLP).[1] The IR (or perhaps more accurately "text retrieval") task may be characterized as the problem of selecting a subset of documents (from a document collection) whose content is relevant to the information need of a user as expressed by a query. The document collections involved in IR are often gigabytes of unrestricted natural-language text. A user's query may be expressed in a controlled language (e.g., a boolean expression of keywords) or, more desirably, a natural language, such as English.

A typical IR system works as follows. The documents to be retrieved are processed to extract *indexing terms* or *content carriers*, which are usually single words or (less typically) phrases. The indexing terms provide a description of the document's content. Weights are often assigned to terms to indicate how well they describe the document. A (natural-language) query is processed in a similar way to extract *query terms*. Query terms are then matched against the indexing terms of a document to determine the relevance of each document to the query.

The ultimate goal of an IR system is to increase both *precision*, the proportion of retrieved documents that are relevant, as well as *recall*, the proportion of relevant document that are retrieved. However, the real challenge is to understand and represent appropriately the content of a document and query, so that the relevance decision can be made efficiently, without degrading precision and recall. A typical solution to the problem of making relevance decisions efficient is to require exact matching of indexing terms and query terms, with an evaluation of the 'hits' based on a scoring metric. Thus, for instance, in vector-space models of relevance ranking, both the indexing terms of a document and the query terms are treated as vectors (with individual term weights) and the similarity between the two vectors is given by a cosine-distance measure, essentially the angle between any two vectors.[2]

### 1.2 Natural-Language Processing for IR

One can regard almost any IR system as performing an NLP task: text is 'parsed' for terms and terms are used to express 'meaning'—to capture document content. Clearly, most traditional IR systems do not attempt to find structure in the natural-language text in the 'parsing' process; they merely extract word-like strings to use in indexing. Ideally, however, extracted structure would directly reflect the encoded linguistic relations among terms—capturing the conceptual content of the text better than simple word-strings.

There are several prerequisites for effective NLP in an IR application, including the following.

---

[1] (Evans, 1990; Evans et al., 1993; Smeaton, 1992; Lewis & Sparck Jones, 1996)

[2] (Salton & McGill, 1983)

1. Ability to process large amounts of text

    The amount of text in the databases accessed by modern IR systems is typically measured in gigabytes. This requires that the NLP used must be extraordinarily efficient in both its time and space requirements. It would be impractical to use a parser with the speed of one or two sentences per second.

2. Ability to process unrestricted text

    The text database for an IR task is generally unrestricted natural-language text possibly encompassing many different domains and topics. A parser must be able to manage the many kinds of problems one sees in natural-language corpora, including the processing of unknown words, proper names, and unrecognized structures. Often more is required, as when spelling, transcription, or OCR errors occur. Thus, the NLP used must be especially robust.

3. Need for shallow understanding

    While the large amount of unrestricted text makes NLP more difficult for IR, the fact that a deep and complete understanding of the text may not be necessary for IR makes NLP for IR relatively easier than other NLP tasks such as machine translation. The goal of an IR system is essentially to classify documents (as relevant or irrelevant) vis-a-vis a query. Thus, it may suffice to have a shallow and partial representation of the content of documents.

Information retrieval thus poses the genuine challenge of processing large volumes of unrestricted natural-language text but not necessarily at a deep level.

### 1.3 Our Work

This paper reports on our evaluation of the use of simple, yet robust and efficient noun-phrase analysis techniques to enhance phrase-based IR. In particular, we explored an extension of the phrase-based indexing in the CLARIT™ system[3] using a hybrid approach to the extraction of meaningful (continuous or discontinuous) subcompounds from complex noun phrases exploiting both corpus-statistics and linguistic heuristics. Using such subcompounds rather than whole noun phrases as indexing terms helps a phrase-based IR system solve the phrase normalization problem, that is, the problem of matching syntactically different, but semantically similar phrases. The results of our experiments show that both recall and precision are improved by using extracted subcompounds for indexing.

## 2 Phrase-Based Indexing

The selection of appropriate indexing terms is critical to the improvement of both precision and recall in an IR task. The ideal indexing terms would directly represent the concepts in a document. Since 'concepts' are difficult to represent and extract (as well as to define), concept-based indexing is an elusive goal. Virtually all commercial IR systems (with the exception of the CLARIT system) index only on 'words', since the identification of words in texts is typically easier and more efficient than the identification of more complex structures. However, single words are rarely specific enough to support accurate discrimination and their groupings are often accidental. An often cited example is the contrast between "junior college" and "college junior". Word-based indexing cannot distinguish the phrases, though their meanings are quite different. Phrase-based indexing, on the other hand, as a step toward the ideal of concept-based indexing, can address such a case directly.

Indeed, it is interesting to note that the use of phrases as index terms has increased dramatically among the systems that participate in the TREC evaluations.[4] Even relatively traditional word-based systems are exploring the use of multi-word terms by supplementing words with statistical phrases—selected high frequency adjacent word pairs (bigrams). And a few systems, such as CLARIT—which uses simplex noun phrases, attested subphrases, and contained words as index terms—and New York University's TREC system[5]—which uses "head–modifier pairs" derived from identified noun phrases—have demonstrated the practicality and effectiveness of thorough NLP in IR tasks.

The experiences of the CLARIT system are instructive. By using selective NLP to identify simplex NPs, CLARIT generates phrases, subphrases, and individual words to use in indexing documents and queries. Such a first-order analysis of the linguistic structures in texts approximates concepts and affords us alternative methods for calculating the fit between documents and queries. In particular, we can choose to treat some phrasal structures as atomic units and others as additional information about (or representations of) content. There are immediate effects in improving precision:

1. Phrases can replace individual indexing words. For example, if both "dog" and "hot" are used for indexing, they will match any query in which both words occur. But if only the phrase "hot dog" is used as an index term, then it will only match the same phrase, not any of the individual words.

---

[3](Evans et al., 1991; Evans et al., 1993; Evans et al., 1995; Evans et al., 1996)

[4](Harman, 1995; Harman, 1996)

[5](Strzalkowski, 1994)

2. Phrases can supplement word-level matches. For example, if only the individual words "junior" and "college" are used for indexing, both "junior college" and "college junior" will match a query with the phrase "junior college" equally well. But if we also use the phrase "junior college" for indexing, then "junior college" will match better than "college junior", even though the latter also will receive some credit as a match at the word level.

We can see, then, that it is desirable to distinquish—and, if possible, extract—two kinds of phrases: those that behave as *lexical atoms* and those that reflect more general linguistic relations.

Lexical atoms help us by obviating the possibility of extraneous word matches that have nothing to do with true relevance. We do not want "hot" or "dog" to match on "hot dog". In essence, we want to eliminate the effect of the independence assumption at the word level by creating new words—the lexical atoms—in which the individual word dependencies are explicit (structural).

More general phrases help us by adding detail. Indeed, all possible phrases (or paraphrases) of actual content in a document are potentially valuable in indexing. In practice, of course, the indexing term space has to be limited, so it is necessary to select a subset of phrases for indexing. Short phrases (often nominal compounds) are preferred over long complex phrases, because short phrases have better chances for matching short phrases in queries and will still match longer phrases owing to the short phrases they have in common. Using only short phrases also helps solve the phrase normalization problem of matching syntactically different long phrases (when they share similar meaning).[6]

Thus, lexical atoms and small nominal compounds should make good indexing phrases.

While the CLARIT system does index at the level of phrases and subphrases, it does not currently index on lexical atoms or on the small compounds that can be derived from complex NPs, in particular, reflecting cross-simplex NP dependency relations. Thus, for example, under normal CLARIT processing the phrase "the quality of surface of treated stainless steel strip"[7] would yield index terms such as "treated stainless steel strip", "treated stainless steel", "stainless steel strip", and "stainless steel" (as a phrase, not lexical atom), along with all the relevant single-word terms in the phrase. But the process would not identify "stainless steel" as a potential lexical atom or find terms such as "surface quality", "strip surface", and "treated strip".

To achieve more complete (and accurate) phrase-based indexing, we propose to use the following

---
[6](Smeaton, 1992)
[7]This is an actual example from a U.S. patent document.

four kinds of phrases as indexing terms:

1. Lexical atoms (e.g., "hot dog" or perhaps "stainless steel" in the example above)
2. Head modifier pairs (e.g., "treated strip" and "steel strip" in the example above)
3. Subcompounds (e.g., "stainless steel strip" in the example above)
4. Cross-preposition modification pairs (e.g., "surface quality" in the example above)

In effect, we aim to augment CLARIT indexing with lexical atoms and phrases capturing additional (discontinuous) modification relations than those that can be found within simplex NPs.

It is clear that a certain level of robust and efficient noun-phrase analysis is needed to extract the above four kinds of small compounds from a large unrestricted corpus. In fact, the set of small compounds extracted from a noun phrase can be regarded as a weak representation of the meaning of the noun phrase, since each meaningful small compound captures a part of the meaning of the noun phrase. In this sense, extraction of such small compounds is a step toward a shallow interpretation of noun phrases. Such weak interpretation is useful for tasks like information retrieval, document classification, and thesaurus extraction, and indeed forms the basis in the CLARIT system for automated thesaurus discovery.

## 3 Methodology

Our task is to parse text into NPs, analyze the noun phrases, and extract the four kinds of small compounds given above. Our emphasis is on robust and efficient NLP techniques to support large-scale applications.

For our purposes, we need to be able to identify all simplex and complex NPs in a text. Complex NPs are defined as a sequence of simplex NPs that are associated with one another via prepositional phrases. We do not consider simplex NPs joined by relative clauses.

Our approach to NLP involves a hybrid use of corpus statistics supplemented by linguistic heuristics. We assume that there is no training data (making the approach more practically useful) and, thus, rely only on statistical information in the document database itself. This is different from many current statistical NLP techniques that require a training corpus. The volume of data we see in IR tasks also makes it impractical to use sophisticated statistical computations.

The use of linguistic heuristics can assist statistical analysis in several ways. First, it can focus the use of statistics by helping to eliminate irrelevant structures from consideration. For example, syntactic category analysis can filter out impossible word modification pairs, such as [adjective, adjective] and [noun, adjective]. Second, it may improve the reliability of statistical decisions. For example, the counting of bigrams that occur only within noun phrases is more reliable for lexical atom discovery than the counting of all possible bigrams that occur in the corpus. In addition, syntactic category analysis is also helpful in adjusting cutoff parameters for statistics. For example, one useful heuristic is that we should use a higher threshold of reliability (evidence) for accepting the pair [adjective, noun] as a lexical atom than for the pair [noun, noun]: a noun–noun pair is much more likely to be a lexical atom than an adjective–noun one.

The general process of phrase generation is illustrated in Figure 1. We used the CLARIT NLP module as a preprocessor to produce NPs with syntactic categories attached to words. We did not attempt to utilize CLARIT complex-NP generation or subphrase analysis, since we wanted to focus on the specific techniques for subphrase discovery that we describe in this paper.

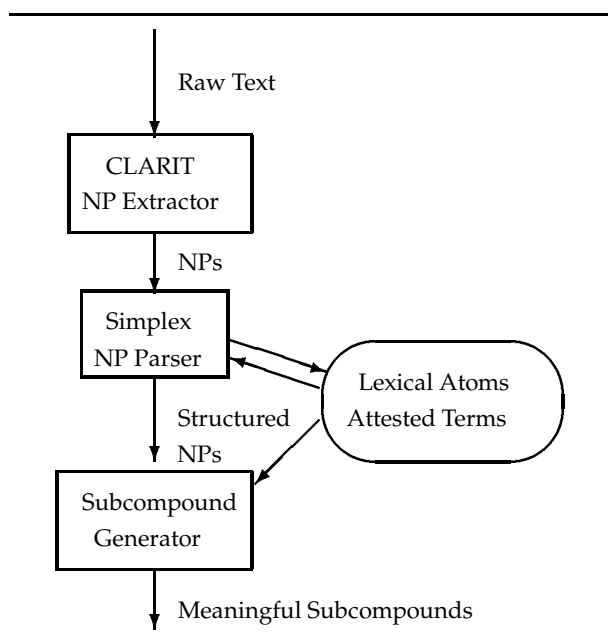

Figure 1: General Processing for Phrase Generation

After preprocessing, the system works in two stages—parsing and generation. In the parsing stage, each simplex noun phrase in the corpus is parsed. In the generation stage, the structured noun phrase is used to generate candidates for all four kinds of small compounds, which are further tested for occurrence (validity) in the corpus.

Parsing of simplex noun phrases is done in multiple phases. At each phase, noun phrases are partially parsed, then the partially parsed structures are used as input to start another phase of partial parsing. Each phase of partial parsing is completed by concatenating those most reliable modification pairs together to form a single unit. The reliability of a modification pair is determined by a score based on frequency statistics and category analysis and is further tested via local optimum phrase analysis (described below). Lexical atoms are discovered at the same time, during simplex noun phrase parsing.

Phrase generation is quite simple. Once the structure of a noun phrase (with marked lexical atoms) is known, the four kinds of small compounds can be easily produced. Lexical atoms are already available. Head–modifier pairs can be extracted based on the modification relations implied by the structure. Subcompounds are just the substructures of the NP. Cross-preposition pairs are generated by enumerating all possible pairs of the heads of each simplex NP within a complex NP in backward order.[8]

To validate discontinuous compounds such as non-sequential head–modifier pairs and cross-preposition pairs, we use a standard technique of CLARIT processing, viz., we test any nominated compounds against the corpus itself. If we find independently attested (whole) simplex NPs that match the candidate compounds, we accept the candidates as index terms. Thus for the NP "the quality of surface of treated stainless steel strip", the head–modifier pairs "treated strip", "stainless steel", "stainless strip", and "steel strip", and the cross-preposition pairs "strip surface", "surface quality", and "strip quality", would be generated as index terms only if we found independent evidence of such phrases in the corpus in the form of free-standing simplex NPs.

### 3.1 Lexical Atom Discovery

A lexical atom is a semantically coherent phrase unit. Lexical atoms may be found among proper names, idioms, and many noun–noun compounds. Usually they are two-word phrases, but sometimes they can consist of three or even more words, as in the case of proper names and technical terms. Examples of lexical atoms (in general English) are "hot dog", "tear gas", "part of speech", and "von Neumann".

However, recognition of lexical atoms in free text is difficult. In particular, the relevant lexical atoms for a corpus of text will reflect the various discourse domains encompassed by the text. In a collection of medical documents, for example, "Wilson's disease" (an actual rheumatological disorder) may be used as a lexical atom, whereas in a collection of general news stories, "Wilson's disease" (reference to the disease that Wilson has) may not be a lexical atom. Note that in the case of the medical usage, we would commonly find "Wilson's disease" as a bigram and we would not find, for example,

---

[8](Schwarz, 1990) reports a similar strategy.

"Wilson's severe disease" as a phrase, though the latter might well occur in the general news corpus. This example serves to illustrate the essential observation that motivates our heuristics for identifying lexical atoms in a corpus: (1) words in lexical atoms have strong association, and thus tend to co-occur as a phrase and (2) when the words in a lexical atom co-occur in a noun phrase, they are never or rarely separated.

The detection of lexical atoms, like the parsing of simplex noun phrases, is also done in multiple phases. At each phase, only two adjacent units are considered. So, initially, only two-word lexical atoms can be detected. But, once a pair is determined to be a lexical atom, it will behave exactly like a single word in subsequent processing, so, in later phases, atoms with more than two words can be detected.

Suppose the pair to test is $[W_1, W_2]$. The first heuristic is implemented by requiring the frequency of the pair to be higher than the frequency of any other pair that is formed by either word with other words in common contexts (within a simplex noun phrase). The intuition behind the test is that (1) in general, the high frequency of a bigram in a simple noun phrase indicates strong association and (2) we want to avoid the case where $[W_1, W_2]$ has a high frequency, but $[W_1, W_2, W]$ (or $[W, W_1, W_2]$) has an even higher frequency, which implies that $W_2$ (or $W_1$) has a stronger association with $W$ than with $W_1$ (or $W_2$, respectively). More precisely, we require the following:

$F(W_1, W_2) > MaxLDF(W_1, W_2)$

and

$F(W_1, W_2) > MaxRDF(W_1, W_2)$

Where,

$MaxLDF(W_1, W_2) =$
$Max_W(Min(F(W, W_1), DF(W, W_2)))$

and

$MaxRDF(W_1, W_2) =$
$Max_W(Min(DF(W_1, W), F(W_2, W)))$

$W$ is any context word in a noun phrase and $F(X, Y)$ and $DF(X, Y)$ are the continuous and discontinuous frequencies of $[X, Y]$, respectively, within a simple noun phrase, i.e., the frequency of patterns $[...X, Y...]$ and patterns $[...X, ..., Y...]$, respectively.

The second heuristic requires that we record all cases where two words occur in simplex NPs and compare the number of times the words occur as a strictly adjacent pair with the number of times they are separated. The second heuristic is simply implemented by requiring that $F(W_1, W_2)$ be much higher than $DF(W_1, W_2)$ (where 'higher' is determined by some threshold).

Syntactic category analysis also helps filter out impossible lexical atoms and establish the threshold for passing the second test. Only the following category combinations are allowed for lexical atoms: [noun, noun], [noun, lexatom], [lexatom, noun], [adjective, noun], and [adjective, lexatom], where "lexatom" is the category for a detected lexical atom. For combinations other than [noun, noun], the threshold for passing the second test is high.

In practice, the process effectively nominates phrases that are true atomic concepts (in a particular domain of discourse) or are being used so consistently as unit concepts that they can be safely taken to be lexical atoms. For example, the lexical atoms extracted by this process from the CACM corpus (about 1 MB) include "operating system", "data structure", "decision table", "data base", "real time", "natural language", "on line", "least squares", "numerical integration", and "finite state automaton", among others.

### 3.2 Bottom-Up Association-Based Parsing

Extended simplex noun-phrase parsing as developed in the CLARIT system, which we exploit in our process, works in multiple phases. At each phase, the corpus is parsed using the most specific (i.e., recently created) lexicon of lexical atoms. New lexical atoms (results) are added to the lexicon and are reused as input to start another phase of parsing until a complete parse is obtained for all the noun phrases.

The idea of association-based parsing is that by grouping words together (based on association) many times, we will eventually discover the most restrictive (and informative) structure of a noun phrase. For example, if we have evidence from the corpus that "high performance" is a more reliable association and "general purpose" a less reliable one, then the noun phrase "general purpose high performance computer" (an actual example from the CACM corpus) would undergo the following grouping process:

general purpose high performance computer ⇒
general purpose [high=performance] computer ⇒
[general=purpose] [high=performance] computer ⇒
[general=purpose] [[high=performance]=computer] ⇒
[[general=purpose]=[[high=performance]=computer]]

Word pairs are given an association score ($S$) according to the following rules. Scores provide evidence for groupings in our parsing process. Note that a smaller score means a stronger association.

1. Lexical atoms are given score 0. This gives the highest priority to lexical atoms.

2. The combination of an adverb with an adjective, past participle, or progressive verb is given score 0.

3. Syntactically impossible pairs are given score 100. This assigns the lowest priority to those

pairs filtered out by syntactic category analysis. The 'impossible' combinations include pairs such as [noun, adjective], [noun, adverb], [adjective, adjective], [past-participle, adjective], [past-participle, adverb], and [past-participle, past-participle], among others.

4. Other pairs are scored according to the formulas given in Figure 2. Note the following effects of the formulas:
When $F(W_1, W_2)$ increases, $S(W_1, W_2)$ decreases;
When $DF(W_1, W_2)$ increases, $S(W_1, W_2)$ decreases;
When $AvgLDF(W_1, W_2)$ or $AvgRDF(W_1, W_2)$ increases, $S(W_1, W_2)$ increases; and
When $F(W_1) - F(W_1, W_2)$ or $F(W_2) - F(W_1, W_2)$ increases, $S(W_1, W_2)$ decreases.

$$S(W_1, W_2) = \frac{1 + LDF(W_1,W_2) + RDF(W_1,W_2)}{\lambda_1 \times F(W_1,W_2) + DF(W_1,W_2)} \times A(W_1, W_2)$$

$$AvgLDF(W_1, W_2) = \frac{\sum_{W \in LD} Min(F(W,W_1), DF(W,W_2))}{|LD|}$$

$$AvgRDF(W_1, W_2) = \frac{\sum_{W \in RD} Min(F(W_2,W), DF(W_1,W))}{|RD|}$$

$$A(W_1, W_2) = \frac{\lambda_2}{F(W_1) + F(W_2) - 2 \times F(W_1,W_2) + \lambda_2}$$

Where
- $F(W)$ is frequency of word W
- $F(W_1, W_2)$ is frequency of adjacent bigram [$W_1$,$W_2$] (i.e., ...$W_1$ $W_2$ ...)
- $DF(W_1, W_2)$ is frequency of discontinuous bigram [$W_1$,$W_2$] (i.e., ...$W_1$...$W_2$...)
- $LD$ is all left dependents, i.e., $\{W \mid min(F(W, W_1), DF(W, W_2)) \neq 0\}$
- $RD$ is all right dependents, i.e., $\{W \mid min(DF(W_1, W), F(W_2, W)) \neq 0\}$
- $\lambda_1$ is the parameter indicating the relative contribution of $F(W_1, W_2)$ to the score (e.g., 5 in the actual experiment)
- $\lambda_2$ is the parameter to control the contribution of word frequency (e.g., 1000 in the actual experiment)

Figure 2: Formulas for Scoring

The association score (based principally on frequency) can sometimes be unreliable. For example, if the phrase "computer aided design" occurs frequently in a corpus, "aided design" may be judged a good association pair, even though "computer aided" might be a better pair. A problem may arise when processing a phrase such as "program aided design": if "program aided" does not occur frequently in the corpus and we use frequency as the principal statistic, we may (incorrectly) be led to parse the phrase as "[program (aided design)]".

One solution to such a problem is to recompute the bigram occurrence statistics after making each round of preferred associations. Thus, using the example above, if we first make the association "computer aided" everywhere it occurs, many instances of "aided design" will be removed from the corpus. Upon recalculation of the (free) bigram statistics, "aided design" will be demoted in value and the false evidence for "aided design" as a preferred association in some contexts will be eliminated.

The actual implementation of such a scheme requires multiple passes over the corpus to generate phrases. The first phrases chosen must always be the most reliable. To aid us in making such decisions we have developed a metric for scoring preferred associations in their local NP contexts.

To establish a preference metric, we use two statistics: (1) the frequency of the pair in the corpus, $F(W_1, W_2)$, and (2) the number of the times that the pair is *locally dominant* in any NP in which the pair occurs. A pair is locally dominant in an NP iff it has a higher association score than either of the pairs that can be formed from contiguous other words in the NP. For example, in an NP with the sequence $[X, Y, Z]$, we compare $S(X, Y)$ with $S(Y, Z)$; whichever is higher is locally dominant. The preference score ($PS$) for a pair is determined by the ratio of its local dominance count ($LDC$)—the total number of cases in which the pair is locally dominant—to its frequency:

$$PS(W_1, W_2) = \frac{LDC(W_1, W_2)}{F(W_1, W_2)}$$

By definition all two-word NPs score their pairs as locally dominant.

In general, in each processing phase we make only those associations in the corpus where a pair's $PS$ is above a specified threshold. If more than one association is possible (above theshold) in a particular NP, we make all possible associations, but in order of $PS$: the first grouping goes to the pair with highest $PS$, and so on. In practice, we have used 0.7 as the threshold for most processing phases.[9]

## 4 Experiment

We tested the phrase extraction system (PES) by using it to index documents in an actual retrieval task. In particular, we substituted the PES for the default NLP module in the CLARIT system and then indexed a large corpus using the terms nominated by the PES, essentially the extracted small compounds and single words (but not words within a lexical atom). All other normal CLARIT processing—weighting of terms, division of documents into subdocuments (passages), vector-space modeling, etc.—was used in its default mode. As a baseline for comparison, we used standard CLARIT processing of the same corpus, with the NLP module set to return full NPs and their contained words (and no further subphrase analysis).[10]

---
[9] When the phrase data becomes sparse, e.g., after six or seven iterations of processing, it is desirable to reduce the threshold.

The corpus used is a 240-megabyte collection of Associated Press newswire stories from 1989 (AP89), taken from the set of TREC corpora. There are about 3-million simplex NPs in the corpus and about 1.5-million complex NPs. For evaluation, we used TREC queries 51–100,[11] each of which is a relatively long description of an information need. Queries were processed by the PES and normal CLARIT NLP modules, respectively, to generate query terms, which were then used for CLARIT retrieval.

To quantify the effects of PES processing, we used the standard IR evaluation measures of recall and precision. Recall measures how many of the relevant documents have been actually retrieved. Precision measures how many of the retrieved documents are indeed relevant. For example, if the total number of relevant documents is $N$ and the system returns $M$ documents of which $K$ are relevant, then,

$Recall = \frac{K}{N}$

and

$Precision = \frac{K}{M}$.

We used the judged-relevant documents from the TREC evaluations as the gold standard in scoring the performance of the two processes.

## 5 Results

The results of the experiment are given in Tables 1, 2, and 3. In general, we see improvement in both recall and precision.

Recall improves slightly (about 1%), as shown in Table 1. While the actual improvement is not significant for the run of fifty queries, the increase in absolute numbers of relevant documents returned indicates that the small compounds supported better matches in some cases.

Interpolated precision improves significantly, as shown in Table 2. The general improvement in precision indicates that small compounds provide more accurate (and effective) indexing terms than full NPs.

Precision improves at various returned-document levels, as well, as shown in Table 3. Initial precision, in particular, improves significantly. This suggests that the PES could be used to support other IR enhancements, such as automatic feedback of the top-returned documents to expand the initial query for a second retrieval step.[12]

| CLARIT | Retrieved-Rel | Total-Rel | Recall |
|---|---|---|---|
| Baseline | 2,668 | 3,304 | 80.8% |
| PES | 2,695 | 3,304 | 81.6% |

Table 1: Recall Results

| Recall | Baseline | PES | Rel.Improvement |
|---|---|---|---|
| 0.00 | 0.6819 | 0.7099 | 4% |
| 0.10 | 0.5535 | 0.5730 | 3.5% |
| 0.20 | 0.4626 | 0.4927 | 6.5% |
| 0.30 | 0.4098 | 0.4329 | 5.6% |
| 0.40 | 0.3524 | 0.3782 | 7.0% |
| 0.50 | 0.3289 | 0.3317 | 0.5% |
| 0.60 | 0.2999 | 0.3026 | 0.9% |
| 0.70 | 0.2481 | 0.2458 | –0.9% |
| 0.80 | 0.1860 | 0.1966 | 5.7% |
| 0.90 | 0.1190 | 0.1448 | 21.7% |
| 1.00 | 0.0688 | 0.0653 | –5.0% |

Table 2: Interpolated Precision Results

| Doc-Level | Baseline | PES | Rel.Improvement |
|---|---|---|---|
| 5 docs | 0.4255 | 0.4809 | 13% |
| 10 docs | 0.4170 | 0.4426 | 6% |
| 15 docs | 0.3943 | 0.4227 | 7% |
| 20 docs | 0.3819 | 0.3957 | 4% |
| 30 docs | 0.3539 | 0.3603 | 2% |
| 100 docs | 0.2526 | 0.2553 | 1% |
| 200 docs | 0.1770 | 0.1844 | 4% |
| 500 docs | 0.0973 | 0.0994 | 2% |
| 1000 docs | 0.0568 | 0.0573 | 1% |

Table 3: Precision at Various Document Levels

The PES, which was not optimized for processing, required approximately 3.5 hours per 20-megabyte subset of AP89 on a 133-MHz DEC alpha processor.[13] Most processing time (more than 2 of every 3.5 hours) was spent on simplex NP parsing. Such speed might be acceptable in some, smaller-scale IR applications, but it is considerably slower than the baseline speed of CLARIT noun-phrase identification (viz., 200 megabytes per hour on a 100-MIPS processor).

---

[10] Note that the CLARIT process used as a baseline does not reflect optimum CLARIT performance, e.g., as obtained in actual TREC evaluations, since we did not use a variety of standard CLARIT techniques that significantly improve performance, such as automatic query expansion, distractor space generation, subterm indexing, or differential query-term weighting. Cf. (Evans et al., 1996) for details.

[11] (Harman, 1993)

[12] (Evans et al., 1995; Evans et al., 1996)

[13] Note that the machine was not dedicated to the PES processing; other processes were running simultaneously.

## 6 Conclusions

The notion of association-based parsing dates at least from (Marcus, 1980) and has been explored again recently by a number of researchers.[14] The method we have developed differs from previous work in that it uses linguistic heuristics and locality scoring along with corpus statistics to generate phrase associations.

The experiment contrasting the PES with baseline processing in a commercial IR system demonstrates a direct, positive effect of the use of lexical atoms, subphrases, and other pharase associations across simplex NPs. We believe the use of NP-substructure analysis can lead to more effective information management, including more precise IR, text summarization, and concept clustering. Our future work will explore such applications of the techniques we have described in this paper.

## 7 Acknowledgements

We received helpful comments from Bob Carpenter, Christopher Manning, Xiang Tong, and Steve Handerson, who also provided us with a hash table manager that made the implementation easier. The evaluation of the experimental results would have been impossible without the help of Robert Lefferts and Nataša Milić-Frayling at CLARITECH Corporation. Finally, we thank the anonymous reviewers for their useful comments.

## References

David A. Evans. 1990. Concept management in text via natural-language processing: The CLARIT approach. In: *Working Notes of the 1990 AAAI Symposium on "Text-Based Intelligent Systems"*, Stanford University, March, 27–29, 1990, 93–95.

David A. Evans, Kimberly Ginther-Webster, Mary Hart, Robert G. Lefferts, Ira A. Monarch. 1991. Automatic indexing using selective NLP and first-order thesauri. In: A. Lichnerowicz (ed.), *Intelligent Text and Image Handling. Proceedings of a Conference, RIAO '91*. Amsterdam, NL: Elsevier, pp. 624–644.

David A. Evans, Robert G. Lefferts, Gregory Grefenstette, Steven K. Handerson, William R. Hersh, and Armar A. Archbold. 1993. CLARIT TREC design, experiments, and results. In: Donna K. Harman (ed.), *The First Text REtrieval Conference (TREC-1)*. NIST Special Publication 500-207. Washington, DC: U.S. Government Printing Office, pp. 251–286; 494–501.

David A. Evans, and Robert G. Lefferts. 1995. CLARIT–TREC experiments *Information Processing and Management*, Vol. 31, No. 3, 385–395.

David A. Evans, Nataša Milić-Frayling, Robert G. Lefferts. 1996. CLARIT TREC-4 experiments. In: Donna K. Harman (ed.), *The Fourth Text REtrieval Conference (TREC-4)*. NIST Special Publication. Washington, DC: U.S. Government Printing Office.

Donna K. Harman, ed. 1993. *The First Text REtrieval Conference (TREC-1)* NIST Special Publication 500-207. Washington, DC: U.S. Government Printing Office.

Donna K. Harman, ed. 1995. *Overview of the Third Text REtrieval Conference (TREC-3)*, NIST Special Publication 500-225. Washington, DC: U.S. Government Printing Office.

Donna K. Harman, ed. 1996. *Overview of the Fourth Text REtrieval Conference (TREC-4)*, NIST Special Publication. Washington, DC: U.S. Government Printing Office.

Mark Lauer. 1995. Corpus statistics meet with the noun compound: Some empirical results. In: *Proceedings of the 33th Annual Meeting of the Association for Computational Linguistics*.

David Lewis and K. Sparck Jones. 1996. Natural language processing for information retrieval. *Communications of the ACM*, January, Vol. 39, No. 1, 92–101.

Mark Liberman and Richard Sproat. 1992. The stress and structure of modified noun phrases in English. In: I. Sag and A. Szabolcsi (eds.), *Lexical Matters*, CSLI Lecture Notes No. 24. Chicago, IL: University of Chicago Press, pp. 131–181.

Mitchell Maucus. 1980. *A Theory of Syntactic Recognition for Natural Language*. Cambridge, MA: MIT Press.

J. Pustejovsky, S. Bergler, and P. Anick. 1993. Lexical semantic techniques for corpus analysis. In: *Computational Linguistics*, Vol. 19(2), Special Issue on Using Large Corpora II, pp. 331–358.

P. Resnik, and M. Hearst. 1993. Structural Ambiguity and Conceptual Relations. In: *Proceedings of the Workshop on Very Large Corpora: Academic and Industrial Perspectives*, June 22, Ohio State University, pp. 58–64.

Gerard Salton and Michael McGill. 1983. *Introduction to Modern Information Retrieval*, New York, NY: McGraw-Hill.

Christoph Schwarz. 1990. Content based text handling. *Information Processing and Management*, Vol. 26(2), pp. 219–226.

Alan F. Smeaton. 1992. Progress in application of natural language processing to information retrieval. *The Computer Journal*, Vol. 35, No. 3, pp. 268–278.

T. Strzalkowski and J. Carballo. 1994. Recent developments in natural language text retrieval. In: Donna K. Harman (ed.), *The Second Text REtrieval Conference (TREC-2)*. NIST Special Publication 500-215. Washington, DC: U.S. Government Printing Office, pp. 123–136.

---

[14](Liberman et al., 1992; Pustejovsky et al., 1993; Resnik et al., 1993; Lauer, 1995)